\pdfoutput=1

\documentclass[11pt]{article}

\usepackage[preprint]{acl}

\usepackage{times}
\usepackage{latexsym}

\usepackage[T1]{fontenc}

\usepackage[utf8]{inputenc}

\usepackage{microtype}

\usepackage{inconsolata}

\usepackage{graphicx}

\usepackage{booktabs}
\usepackage{multirow}
\usepackage{paralist}
\usepackage{bm}
\usepackage{amsmath}

\newcommand{\ie}{\emph{i.e.,}~}
\newcommand{\eg}{\emph{e.g.,}~}

\title{
Towards Text-Image Interleaved Retrieval
}

\author{
 \textbf{Xin Zhang\textsuperscript{1,2}}\footnotemark[1],
 \textbf{Ziqi Dai\textsuperscript{1}}\footnotemark[1],
 \textbf{Yongqi Li\textsuperscript{2}},
 \textbf{Yanzhao Zhang},
 \textbf{Dingkun Long},
\\
 \textbf{Pengjun Xie},
 \textbf{Meishan Zhang\textsuperscript{1}},
 \textbf{Jun Yu\textsuperscript{1}},
 \textbf{Wenjie Li\textsuperscript{2}},
 \textbf{Min Zhang\textsuperscript{1}}
\\
 \textsuperscript{1}Harbin Institute of Technology, Shenzhen
 \textsuperscript{2}The Hong Kong Polytechnic University
\\
 \small{
 \footnotemark[1]Equal contribution. Will release at \url{https://github.com/vec-ai/wikiHow-TIIR}
 }
}

\begin{document}
\maketitle
\begin{abstract}
Current multimodal information retrieval studies mainly focus on single-image inputs, which limits real-world applications involving multiple images and text-image interleaved content. In this work, we introduce the text-image interleaved retrieval (TIIR) task, where the query and document are interleaved text-image sequences, and the model is required to understand the semantics from the interleaved context for effective retrieval. We construct a TIIR benchmark based on naturally interleaved wikiHow tutorials, where a specific pipeline is designed to generate interleaved queries. To explore the task, we adapt several off-the-shelf retrievers and build a dense baseline by interleaved multimodal large language model (MLLM). We then propose a novel Matryoshka Multimodal Embedder (MME), which compresses the number of visual tokens at different granularity, to address the challenge of excessive visual tokens in MLLM-based TIIR models. Experiments demonstrate that simple adaption of existing models does not consistently yield effective results. Our MME achieves significant improvements over the baseline by substantially fewer visual tokens. We provide extensive analysis and will release the dataset and code to facilitate future research.
\end{abstract}

\section{Introduction}

\begin{figure}[t]
\includegraphics[width=\columnwidth]{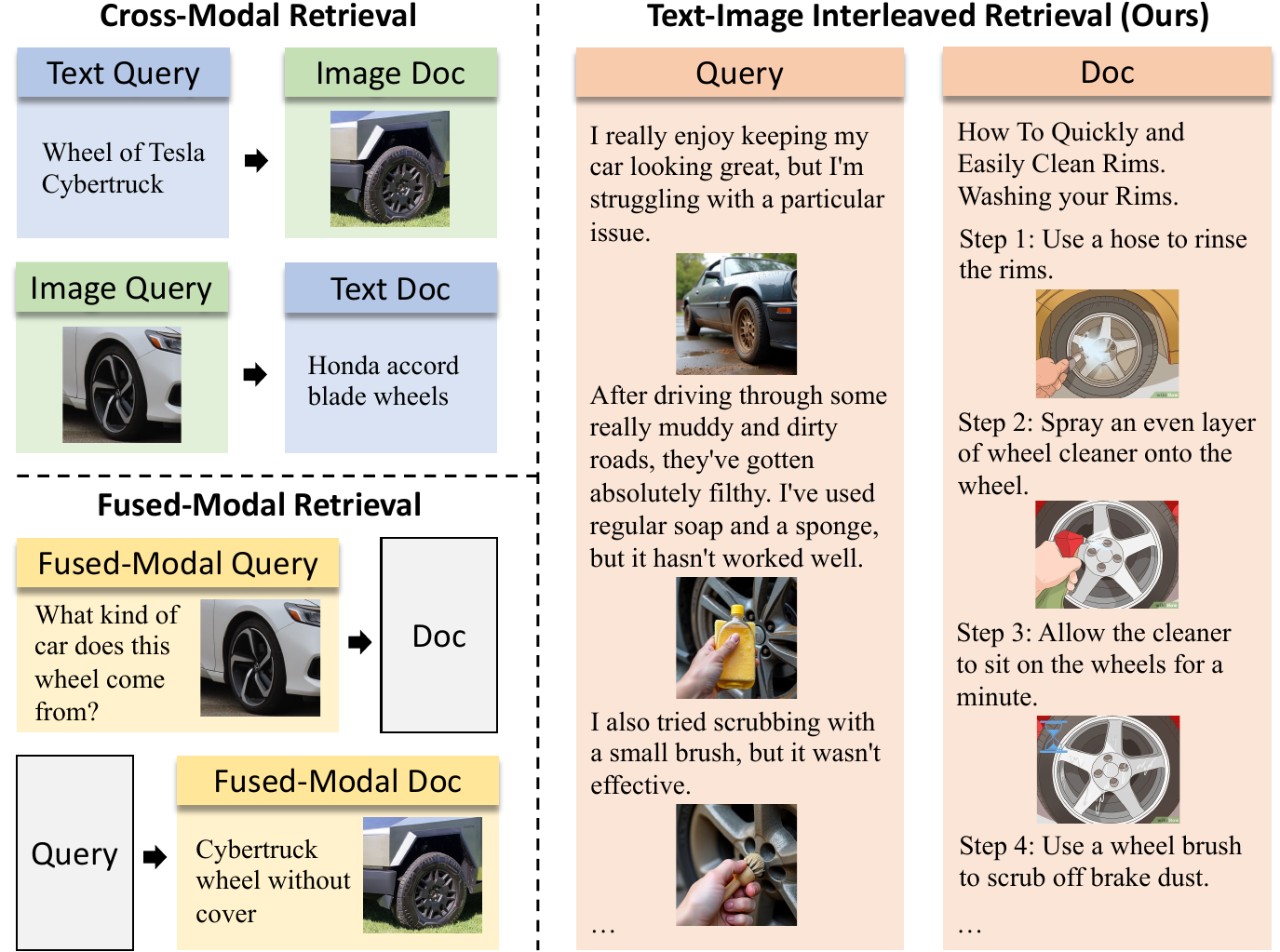}
\caption{
Comparison of our Text-Image Interleaved Retrieval task to previous settings.
Blocks with black borders represent data in text, image or fused-modal.
}
\label{fig:task}
\end{figure}

Multimodal information retrieval (MIR) aims to retrieve relevant information involving multiple modalities \cite{DBLP:conf/eccv/WeiCCHZFRC24}, which plays a crucial role in various applications such as e-commerce search \cite{Wu_2021_CVPR} and retrieval augmented generation (RAG) systems \cite{chen-etal-2022-murag,pmlr-v202-yasunaga23a}.
Current advanced multimodal retrievers \cite{zhou-etal-2024-vista,lin2024mm} typically adopt the dense retrieval paradigm, where queries or documents are encoded into embeddings for vector similarity calculation.
These models have demonstrated promising results in scenarios involving cross-modal and fused-modal retrieval
(Figure \ref{fig:task} left illustrates the settings).

Despite their effectiveness, most existing MIR studies permit only a single image in the query or document \cite{zhou-etal-2024-vista,DBLP:conf/eccv/WeiCCHZFRC24}.
This would largely limit users to clearly present their information needs and requirements, while also restricting the system from leveraging useful documents containing multiple images and interleaved text-image contents.
For example, a tutorial for everyday skills, such as furniture assembly or cooking recipes, typically requires multiple illustrations to describe sequential steps (Figure \ref{fig:task} right).
Similarly, users may need more than one photo to effectively describe their current problems or situations.
Such cases would be inevitable in real-world RAG systems, demonstrating the necessity of interleaved-modal inputs in retrieval.

To explore the above scenarios, we introduce the text-image interleaved retrieval (TIIR) task, where both the query and document contain interleaved text and images (Figure~\ref{fig:task} right).
In TIIR, multiple text chunks and images are sequentially positioned in a semantic manner, allowing for a more accurate expression of user intent and document information.
However, this also presents challenges in understanding interleaved-modal content.

To advance the progress of TIIR, we first construct a new benchmark
based on wikiHow\footnote{
\url{https://www.wikihow.com}.
}, a large-scale collection of human-curated how-to guides with text and images \cite{yang-etal-2021-visual}.
We convert the tutorial articles into a retrieval corpus of 150K interleaved documents.
To obtain interleaved contextual queries, we design a novel and efficient pipeline that leverages powerful large language models (LLMs) \cite{laurenccon2024building,qwen2} and a text-to-image generator \cite{flux2023} to automatically generate interleaved queries (\S\ref{sec:tiir:build}) based on documents.
We then employ human experts to annotate and filter out generation artifacts, resulting in 7,654 high-quality query-document pairs for testing, while the remaining generated queries are allocated to the training set.
We dub this dataset as \texttt{wikiHow-TIIR}.

Beyond the data, building an effective TIIR model is complex due to the challenges in modeling interleaved-modal content.
First, existing retrievers struggle to handle this task effectively due to their single-image constraints.
Second, while fine-tuning multimodal LLMs (MLLMs) with interleaved inputs support \cite{lu2024deepseek-vl} as dense TIIR models seems promising, the hundreds of visual tokens required per image \cite{yin2023survey} leads to prohibitively long sequences, resulting in both computational inefficiency and disproportionate visual dominance in the embedding space (\S\ref{sec:exp:analysis}).
To address these issues, we propose a novel retriever, \ie Matryoshka Multimodal Embedder (MME), that compresses the number of visual tokens at different granularity (\S\ref{sec:model}), thereby generating more effective embeddings for TIIR.

We conduct extensive experiments to explore our dataset and evaluate different retrievers (\S\ref{sec:exp}).
Results show that the interleaved context is the key of TIIR modeling.
Even with specialized adaption strategies, existing retrievers (non-interleaved) perform worse than the native-interleaved baseline, indicating the necessity of developing dedicated TIIR retrievers.
In contrast, our suggested MME outperform the baseline by a large margin, demonstrating the effectiveness of our approach.
We further conduct comprehensive analyses (\S\ref{sec:exp:analysis}) to understand the TIIR task and models.

Our contributions are as follows:
\begin{compactitem}
\item We identify the text-image interleaved retrieval (TIIR) task and construct the wikiHow-TIIR benchmark. To the best of our knowledge, it is the first dataset for TIIR.
\item We propose a novel TIIR model that compresses the number of visual tokens at different granularity, which successfully addresses the challenge in modeling interleaved content.
\item We present extensive experiments and analyses on our dataset, including strategies for adapting existing retrievers. This provides insights for future work and applications.
\end{compactitem}

\section{WikiHow-TIIR Benchmark}\label{sec:tiir}

\subsection{Task Definition}\label{sec:tiir:def}
We first define the text-image interleaved data instance $X$ as a sequence of text and images, $X = [x_i, \cdots, x_n ]$, where $x_i$ can be either a text chunk or an image, all of which are ordered contextually.
Given a query $X^Q$ and a corpus $C$ consisting of documents $\{ X^D_1, \cdots, X^D_m \}$, the TIIR task is to retrieve the most relevant document $X^D$ from $C$ for $X^Q$.
The relevance is determined by a similarity function $f(X^Q, X^D)$, which measures the semantic similarity at the image-text sequence level.
The model is required to understand the semantics from contextually interleaved text and images for effective retrieval, which could be challenging to existing multimodal retrievers.

\subsection{Data Construction}\label{sec:tiir:build}

One of the most common scenarios involving interleaved text and images in everyday life is found in tutorials for daily skills or product manuals, where images are necessary to provide clearer and more vivid descriptions.
WikiHow\footnotemark[1] is a widely used tutorial website that contains a large number of high-quality text-image tutorials that meet these criteria.
Therefore, we choose wikiHow articles crawled by \citet{yang-etal-2021-visual} as our \emph{retrieval corpus}.
For each tutorial, we select the goal, step titles and corresponding images to build an interleaved document.
We then generate and annotate queries.

\begin{figure*}[t]\centering
\includegraphics[width=\linewidth,trim={0 5pt 0 2pt},clip]{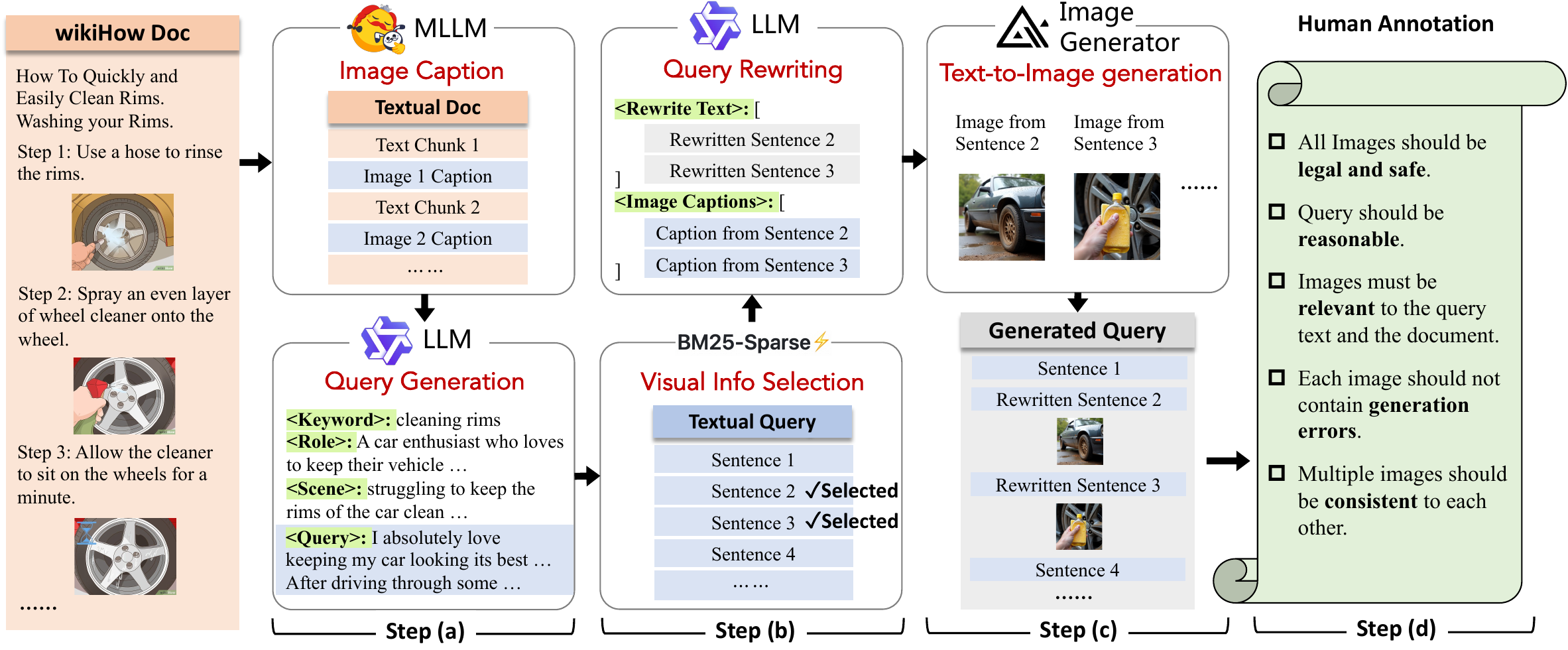}
\caption{
Our data construction workflow (\S\ref{sec:tiir:build}), where step (a), (b) and (c) comprise the generation pipeline, and (d) shows the brief annotation guideline.
Technical details and principles are provided in Appendix \ref{sec:app:data:query-gen} and \ref{sec:app:data:annotation}.
}\label{fig:query-gen}
\end{figure*}

\paragraph{Query Generation}
We design a query generation pipeline to mimic real-world scenarios where users may provide multiple images and text to describe their problems or situations.
Given that current interleaved MLLMs are not yet sufficiently capable of handling complex text and image generation, our pipeline centers on the text modality.
It leverages image caption and text-to-image generation for modality conversion, while utilizing more advanced LLMs to drive query text generation.
As shown in Figure~\ref{fig:query-gen}, it consists of three stages:

\noindent\textbf{(a)} \textit{Query text generation}. 
Given a interleaved document $X^D$, we first generate caption for each image by a strong and efficient MLLM\footnote{
\url{hf.co/HuggingFaceM4/Idefics3-8B-Llama3}
} \cite{laurenccon2024building}.
Then,
based on the tutorial text and image captions,
we instruct a powerful LLM\footnote{
\url{hf.co/Qwen/Qwen2.5-72B-Instruct}
} \cite{qwen2} to write a text query $T^Q$ target to one specific step of the document.

\noindent\textbf{(b)} \textit{Text-image information reorganization}.
We split the query text into sentences and employ BM25 \cite{robertson2009probabilistic} to identify the most informative ones $S_\text{top-k}$ for transforming the textual information into images.
Next, we use the LLM to select entities or actions from $S_\text{top-k}$ to generate captions $C^Q$ for images in query and rewrite the query text into $T^Q_r$ to remove selected information.

\noindent\textbf{(c)} \textit{Image generation}. We use a text-to-image generator\footnote{
\url{hf.co/black-forest-labs/FLUX.1-dev}
} \cite{flux2023} to generate images from image captions $C^Q$ and merge with the rewrited query $T^Q_r$ to form the final query $X^Q$.

We select around 80.7k tutorials from the corpus and generate one query for each tutorial.
As the generated query is designed to be relevant to the corresponding tutorial, we take the tutorial as the \emph{positive} document for the query.

\begin{table}
\centering
\setlength{\tabcolsep}{3pt}
\resizebox{\columnwidth}{!}{
\begin{tabular}{lrccc}
\toprule
\multirow{2}{*}{Part} & \multirow{2}{*}{\#Examples} & Avg./Min/Max & Avg. Text & \multirow{2}{*}{\#Pos.} \\
& & \#Images &  \#Tokens \\ \midrule
Corpus       &  155,262  & 4.97 / 2 / 64 &   85.62 & - \\ \midrule
Train Query  &   73,084  & 2.88 / 2 / 4~ &  105.15 & 1 \\
Test Query   &    7,654  & 2.81 / 2 / 4~ &  105.59 & 1 \\ \bottomrule
\end{tabular}}
\caption{
Statistics of our constructed wikiHow-TIIR dataset, where \texttt{Pos.}\ denotes positive document.
We count tokens by LLaMA tokenizer.
} \label{tab:data-stats}
\end{table}

\paragraph{Testset Annotation}
To build a high-quality testset for fair and reasonable evaluation, we further conduct a human annotation process to filter out generation artifacts and ensure the generated queries are reasonable and contextually interleaved.
Our annotation guidelines primarily focus on five types of issues:
\textbf{(1)} Images must not involve illegal content, sensitive topics, or contain offensive material such as pornography.
\textbf{(2)} The overall content of the query should be reasonable and consistent with common sense.
\textbf{(3)} Images must be relevant to the query text and the document.
\textbf{(4)} Each image in the query should not contain obvious structural or textual errors.
\textbf{(5)} If multiple images in the query depict the same subject or scene, they should not exhibit significant variations.
We select around 10,000 query-document pairs with diverse wikiHow topic labels for annotation, resulting in 7,654 high-quality pairs as the final testset.

\subsection{Data Statistics}\label{sec:tiir:stat}

Table \ref{tab:data-stats} shows the statistics of the \texttt{wikiHow-TIIR} dataset.
From all generated queries, we annotate 7,654 query-positive pairs as the testset, and the remaining 73,084 pairs are used as the trainset.
We present a pie chart of the testset content categories (\eg Food, Pets, Sports) in Appendix Figure \ref{fig:test-category}.

\begin{figure*}[t]\centering
\includegraphics[width=\linewidth]{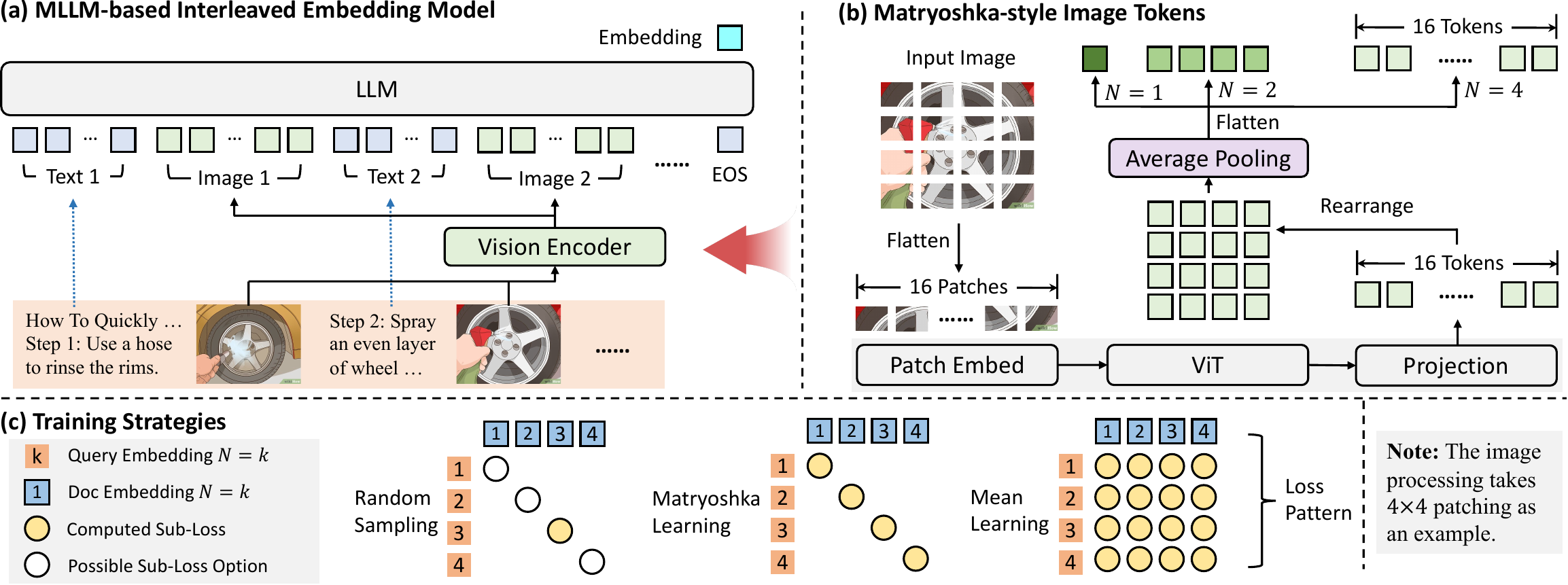}
\caption{
Our TIIR model overview, where (a) is the DPR baseline (\S\ref{sec:method:baseline}), (b) illustrates the computation of visual tokens in different granularities, and (c) shows the training strategies of our MME.
}\label{fig:model}
\end{figure*}

\section{Approach}\label{sec:model}

\subsection{Baseline Model}\label{sec:method:baseline}
Our baseline is in the dense retrieval paradigm, where inputs are encoded by a backbone and a pooling operator is applied to obtain the sequence-level embeddings.
To effectively model the semantics of interleaved context, the interleaved MLLM is a natural backbone choice as the order of text chunks and images are kept in the input sequence and thus attention operations can better capture the multimodal interactions.
In practice, we use the DeepSeek-VL \cite{lu2024deepseek-vl} as the backbone 
and take [EOS] output state as the embedding.

We train it by InfoNCE \cite{oord2018representation} loss:
\begin{equation} \label{eq:infonce}
\mathcal{L} = - \log \frac{ \text{exp}(s(X^Q, X^D_+) / \tau) }{
\sum_{i=1}^{N}\text{exp}(s(X^Q, X^D_i)/ \tau)
}  ,
\end{equation}
where $\tau$ denotes the temperature parameter.
The $X^D_+$ is the relevant document (positive) to $X^Q$, while others are irrelevant documents (negatives), which could be either hard negatives or in-batch negatives.
$s(X^Q, X^D)$ is the relevance score between $X^Q$ and $X^D$, measured by the cosine similarity between their respective embeddings.

\begin{table*}
\resizebox{\textwidth}{!}{
\setlength{\tabcolsep}{4pt}
\begin{tabular}{cclr|ccccc}
\toprule
No. & Setting & Model  & \#Param & Recall@5 & MRR@5 & MRR@10 & nDCG@5 & nDCG@10 \\ \midrule
\multicolumn{9}{c}{\it Non-Interleaved Models} \\ \midrule
1 & \multirow{4}{*}{
\begin{tabular}{c}
Text w/ \\
Merged Image
\end{tabular}
} & VISTA  & 0.21B & 45.06 & 31.95 & 33.73 & 33.14 & 35.22  \\
2 & & GME$_\text{Qwen2-VL-2B}$ &  2.21B & 65.85 & 50.12 & 51.65 & 51.18 & 54.06 \\
3 & & E5-V         &  8.36B  & 62.66 & 46.47 & 48.16 & 47.64 & 50.52 \\
4 & & MM-Embed     &  8.18B  & 68.73 & 52.24 & 53.67 & 53.25 & 56.37 \\
\midrule
5 & \multirow{3}{*}{Text w/ Caption} & BGE-v1.5$_\text{large}$ & 0.34B & 39.66 & 27.54 & 29.14 & 28.58 & 30.56 \\
6 & & GTE-v1.5$_\text{large}$ & 0.43B & 41.44 & 27.74 & 29.56 & 28.94 & 31.14 \\
7 & & GTE-Qwen2-7B & 7.61B & 47.24 & 33.40 & 35.28 & 34.63 & 36.85 \\
\midrule
8 & \multirow{2}{*}{Vector-Fusion} & Jina-CLIP-v2        & 0.87B  & 58.80 & 43.29 & 45.00 & 44.44 & 47.17 \\
9 &  & CLIP$_\text{large}$ Fine-tuned & 0.43B & 69.41 & 53.06 & 54.73 & 54.25 & 57.15 \\
\midrule
\multicolumn{9}{c}{\it Native Interleaved Models (Fine-tuned)} \\
\midrule
10 & \multirow{2}{*}{TIIR} & DPR$_\text{DeepSeek-VL}$     &  \multirow{2}{*}{1.98B}  &  74.79 & 59.43 & 60.87 & 60.49 & 63.28  \\
11 & & \bf MME (Ours) $_{N=3}$  &       & \bf 77.40 & \bf 62.07 & \bf 63.40 & \bf 63.01 & \bf 65.91 \\
\bottomrule
\end{tabular}}
\caption{
Evaluation results on our wikiHow-TIIR.
\texttt{Text w/ Merged Image} denotes the interleaved sequence is concatenated into a text-image pair.
The description of \texttt{Vector-Fusion} is in \S\ref{sec:exp:models}.
} \label{tab:main}
\end{table*}

\subsection{Matryoshka Multimodal Embedder}\label{sec:method:mme}
Current MLLMs utilize Vision Transformers (ViTs) to encode images and a linear projection to convert into visual tokens, which are then concatenated with text tokens to form the input of the LLM backbone.
As most models use a large number of visual tokens (e.g., 576) for each image, a substantial number of images from interleaved data could take excessive visual tokens, leading to great inefficiency and allowing visual information to disproportionately dominate the embedding space.
Moreover, the token sequence will be truncated if it's too long to exceed the model's max context length, which may lose critical semantics for retrieval.
Inspired by \citet{cai2024matryoshka}, we introduce a novel Matryoshka Multimodal Embedder (MME) to address these issues.
MME produces a nested set of visual tokens for each image, which is a Matryoshka doll-like sequence across multiple coarse-to-fine granularities (Figure \ref{fig:model}).
At the inference time, we could set a certain token size to meet the requirement, which would be more flexible and efficient.

Technically, we introduce an average pooling after the visual projection of MLLM to compress the visual tokens into different lengths by different-sized pooling kernels.
We take DeepSeek-VL-1.3B as an example.
Its vision encoder\footnote{
\url{hf.co/timm/ViT-L-16-SigLIP-384}
} divides an image into $24 \times 24$ patches (i.e., 576 in total) and outputs $576$ visual features, which are then projected into visual tokens.
We rearrange the visual tokens into a $24 \times 24$ grid and apply average pooling with kernel size $24/N$ to compress into $N \times N$ grid, resulting in flattened $N^2$ visual tokens.
$N \in \{1, 2, 3, 4, 6, 8, 12, 24\}$.

In training, we propose three strategies to learn the nested visual tokens:
\textbf{(1)} \textit{Random sampling} (Rand): We randomly sample a grid width $N$ for each micro-batch, which is a simple and efficient way for the model to adapt inputs at different levels of granularity.
\textbf{(2)} \textit{Matryoshka learning} (MRL): We train the model with all $M$ kernel sizes simultaneously, where the model is trained with a weighted sum of $M$ losses from different grid sizes.
\textbf{(3)} \textit{Mean learning} (Mean): Similar to \texttt{MRL}, but we additionally compute losses between query and document embeddings of different sizes, the final loss is the mean of all $M \times M$ possible combinations.

\section{Experiments}\label{sec:exp}

\subsection{Evaluated Models}\label{sec:exp:models}
Besides the DPR$_\text{DeepSeek-VL}$ baseline (\S\ref{sec:method:baseline}), we also adapt non-interleaved retrievers for evaluation:

$\bullet$ Single-image multimodal models, \ie VISTA \cite{zhou-etal-2024-vista}, E5-V \cite{jiang2024e5}, MM-Embed \cite{lin2024mm} and GME \cite{zhang2024gme}, where we concatenate multiple images into one (Appendix Figure \ref{fig:single-image-example} shows an example).

$\bullet$ Text models, \ie BGE \cite{xiao2024cpack} and GTE \cite{zhang-etal-2024-mgte}. We evaluate them by replacing images with text captions from a MLLM\footnote{
\url{hf.co/Qwen/Qwen2-VL-2B-Instruct}
} (details refer to Appendix \S\ref{sec:app:text-model}).

$\bullet$ CLIP-style two-stream models, we evaluate the well-trained Jina-CLIP\footnote{
\url{hf.co/jinaai/jina-clip-v2}
} \cite{koukounas2024jinaclipv2} and fine-tuned original CLIP \cite{radford2021learning} by a simple vector-fusion strategy.
Given an input sequence, we concatenate all text chunks and encode as one text embedding $\bm{t}$, while all images are separately encoded as image embeddings $\{\bm{i}_1, \dots, \bm{i}_n\}$.
The final embedding $\bm{e}$ is the normalized sum of the normalized average embedding of images and the text embedding, \ie $\bm{e} = \text{Norm}(\text{Norm}(\text{Mean}(\bm{i}_1, \dots, \bm{i}_n)) + \bm{t})$.

\subsection{Settings}

\paragraph{Metrics}
We compute Recall@$k$ (the rate that positives are successfully retrieved within the top-$k$ ranked results), MRR@$k$ (Mean Reciprocal Rank, the average of reciprocal ranks of the first positive in the top-$k$) and nDCG@$k$ (normalized Discounted Cumulative Gain, the quality of ranking by considering both the relevance and position of positives within top-$k$) on our testset for evaluation.

\paragraph{Implementation}
We fine-tune OpenAI CLIP\footnote{
\url{hf.co/openai/clip-vit-large-patch14}
} and DeepSeek-VL-1.3B\footnote{
\url{hf.co/deepseek-ai/deepseek-vl-1.3b-base}
}.
We use a batch size of $32$ and a learning rate of $5 \times 10^{-5}$ with a linear warm-up scheduler to train the models for $3$ epochs.
The contrastive learning temperature $\tau$ is set to $0.05$.
We use in-batch negatives and $1$ randomly selected hard negative.
Other details are provided in Appendix \S\ref{sec:app:implementation}.

\subsection{Main Results}

Table \ref{tab:main} presents the results on our wikiHow-TIIR benchmark.
First, we focus on the evaluation of adapted non-interleaved models.
For the single-image multimodal retrievers (setting \texttt{Text w/ Merge Image} in Table \ref{tab:main}), by combining multiple images into one image, they could achieve reasonable performance.
From \texttt{VISTA} to \texttt{GME} and then to \texttt{MM-Embed}, The scaling of the model size could bring consistent improvements.
While \texttt{E5-V} appears to be an outlier with suboptimal performance, this is understandable given that it was trained solely on textual natural language inference data \cite{jiang2024e5}, without exposure to either retrieval or multimodal data.
It is remarkable to observe that foundation MLLMs can demonstrate such comparable performance.
By replacing images with captions (setting \texttt{Text w/ Caption}), the text retrievers at different sizes perform worse than their similar-sized multimodal models, \eg \texttt{BGE} is worse than \texttt{VISTA}.
This is because captions may not fully preserve the visual semantics (as we will analyze in Table \ref{tab:adapt-ablation}).
Regarding two-stream models, the \texttt{vector-fusion} strategy allows well-finetuned \texttt{Jina-CLIP} \cite{koukounas2024jinaclipv2} to be directly adapted, achieving promising performance.

For native interleaved models, we can observe that:
(1) The DPR baseline (row 10) performs better than fine-tuned CLIP (row 9), demonstrating the interleaved modeling provides a more accurate context understanding for TIIR;
(2) Our proposed MME (row 11) further improves the performance by a large margin, indicating the effectiveness of our Matryoshka-style visual token learning.

In summary, all adapted models are underperformed by the native interleaved models, which calls for developing TIIR support in future multimodal retrievers.
It is also worth noting that, to ensure fair comparison to a reasonable extent, we do not fine-tune any off-the-shelf retrievers, and the fine-tuned models are initialized from weak checkpoints (models that have not been trained on any high-quality retrieval data).

\begin{figure}[t]\centering
\includegraphics[width=\linewidth]{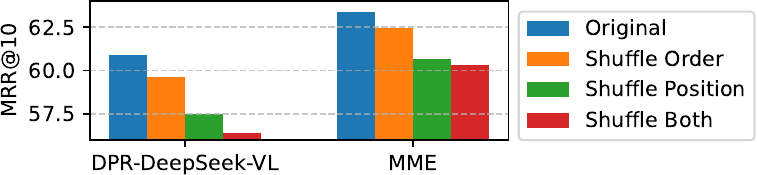}
\caption{
Results of interleaved models evaluated on settings of original data, shuffled image ordering, shuffled image position, and shuffled image ordering \& position.
}\label{fig:simr-img}
\end{figure}

\begin{table}
\resizebox{\linewidth}{!}{
\setlength{\tabcolsep}{4pt}
\begin{tabular}{ccl|cc}
\toprule
\multirow{2}{*}{No.} & Original & \multirow{2}{*}{Model} & \multicolumn{2}{c}{MRR@10} \\
&  Setting &  & Original & Text \\ \midrule
1 & \multirow{4}{*}{
\begin{tabular}{c}
Text w/ \\
Merged Image
\end{tabular}
} & VISTA  &  33.73 & \bf 41.32 \\
2 & & GME$_\text{Qwen2-VL-2B}$ & \bf 51.65 & 43.26 \\
3 & & E5-V         &   \bf 48.16 & 43.76 \\
4 & & MM-Embed     &   53.67 & 53.54 \\
\midrule
5 & \multirow{3}{*}{Text w/ Caption} & BGE-v1.5$_\text{large}$ & 29.14 & \bf 44.55 \\
6 & & GTE-v1.5$_\text{large}$ &  29.56 & \bf 44.35 \\
7 & & GTE-Qwen2-7B &  35.28 & \bf 46.66 \\
\midrule
8 & Vector-Fusion & Jina-CLIP-v2        & \bf 45.00 & 39.78 \\
\midrule
9 & Visual Doc & GME$_\text{Qwen2-VL-2B}$ & \bf 45.92 & 43.26 \\
\bottomrule
\end{tabular}}
\caption{
Comparison of performance between original adaption and text-only evaluation (ignoring images).
The adaption strategy could be considered as useful if text results are lower than the original.
} \label{tab:adapt-ablation}
\end{table}

\subsection{Analysis}\label{sec:exp:analysis}
This subsection presents several in-depth analyses to understand the TIIR task and models.
We address the following five research questions.

\paragraph{RQ1: Can the interleaved context be effectively modeled? Fig. \ref{fig:simr-img}}
Given that text-image interleaved context lies at the core of our task, a natural question arises regarding its importance for retrieval.
We examine this by manipulating the images in several ways:
(1) shuffling the image ordering, (2) shuffling the image position, and (3) shuffling both image ordering and position.
To ensure rigorous evaluation of these settings and isolate other potential confounding factors, we only evaluate the native interleaved models.
Figure \ref{fig:simr-img} demonstrates that shuffling both image ordering and position leads to significant performance degradation, indicating that both the order among images and the alignment between images and text affect the context semantics.
Combining both settings further decreases the result.
In summary, the performance drop empirically demonstrates that the interleaved context is effectively modeled and crucial for accurate retrieval.

\paragraph{RQ2: Are the off-the-shelf models adaptation strategies (\S\ref{sec:exp:models}) effective? Tab. \ref{tab:adapt-ablation}}
After recognizing the importance of interleaved context, we further evaluate the effectiveness of the adaptation strategies (\S\ref{sec:exp:models}) for off-the-shelf models.
A direct probing to this question is hard to achieve, as they are not designed for the TIIR task.
Fortunately, an elegant solution emerges: since all these models are proven to be powerful text retrievers, we could investigate this question by comparing their adapted performance against their text-only retrieval scores.
Table \ref{tab:adapt-ablation} presents the results.
We observe that for single-image multimodal retrievers, the adaption of merging multiple images into one does not always succeed.
We suppose that the merged image (as the example in Figure \ref{fig:single-image-example}) not only loses the interleaved context but also introduces noise in content understanding.
The image caption strategy for text retrievers actually decreases the performance, which could be due to the fact that the generated captions are not as informative as the original images.
Notably, the vector-fusion strategy improves the performance, which could be attributed to the feature-level fusion of text and images.
Nonetheless, we suppose that these failures stem from the loss of interleaved data structure.
Effectively preserving this interleaved context is crucial for enabling existing models to support TIIR.

\paragraph{RQ3: Can we model the interleaved context in the vision modality? Tab. \ref{tab:adapt-ablation}}
All adaptions in \S\ref{sec:exp:models} preserve the original text information.
For vision modality, a promising recent paradigm in multimodal retrieval is based on visual documents \cite{ma-etal-2024-unifying,faysse2024colpali}, which takes screenshots of multimodal documents as input.
Among evaluated models, \texttt{GME} \cite{zhang2024gme} supports this mode.
To explore its potential, we convert interleaved sequences into visual docs (as shown in Appendix Figure \ref{fig:text2image}) for evaluation.
The last row of Table \ref{tab:adapt-ablation} shows the results.
Interestingly, this adaptation is also effective (\ie the adapted scores are higher than that of text-only) as it maintains the interleaved information structure.

\begin{figure}[t]\centering
\includegraphics[width=\linewidth]{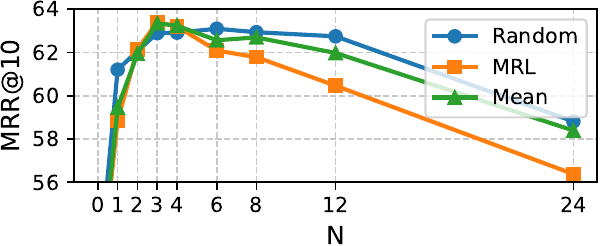}
\caption{
Performance curve of different settings of Matryoshka-style visual token, where all three different training strategies (\S\ref{sec:method:mme}) are presented. The best one (\texttt{mean}) is selected as the final model.
}\label{fig:token-curve}
\end{figure}

\begin{figure}[t]\centering
\includegraphics[width=\linewidth]{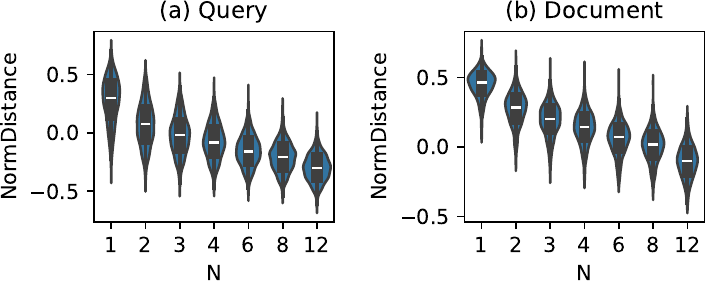}
\caption{
The distribution of the normalization between distances of an embedding with setting $N$ and both text-only embeddings ($d_t$) and full image tokens embeddings ($d_i$), calculated as $(d_i-d_t)/(d_i+d_t)$.
Higher values indicate text information dominance, while lower values suggest stronger visual influence.
The distribution aligns with the performance curve, where the optimal $N=3$ yields a more balanced distribution.
}\label{fig:dis-vis}
\end{figure}

\paragraph{RQ4: Understanding the Matryoshka-style visual token. Fig. \ref{fig:token-curve} \& \ref{fig:dis-vis}}
Now we focus on the proposed MME model.
In Table \ref{tab:main}, for brevity, we only report the results of $N=3$ of the best training strategy.
To better understand the behavior, we display the performance curve of different visual token settings in Figure \ref{fig:token-curve}.
We can see that, for all three training strategies, retrieval performance exhibits an inverted U-shaped relationship with the number of visual tokens, initially improving before declining.
The observed pattern aligns well with the intuition:
an insufficient number of visual tokens fails to capture the rich semantics of each image, while excessive tokens dominate the input sequence, leading to semantic bias in the embeddings as well as inaccurate retrieval results.
This highlights the importance of compressing visual tokens for multiple images and interleaved retrieval models.
In addition, all strategies reach the peak performance at $N=3$, which implies the best visual token size is dataset/domain dependent.
We further investigate the visual information dominance by calculating the normalization between distances of an embedding and both text-only embeddings ($d_t$) and full image tokens embeddings ($d_i$), as $(d_i-d_t)/(d_i+d_t)$, as plotted in Figure \ref{fig:dis-vis}.
The distribution aligns with the performance curve, where the optimal $N=3$ yields a more balanced distribution, indicating a more effective balance between text information and visual influence.

\begin{table}
\resizebox{\linewidth}{!}{
\setlength{\tabcolsep}{4pt}
\begin{tabular}{lcrc}
\toprule
\multirow{2}{*}{Setting} & Avg. Seq. Len. & \multirow{2}{*}{Encoding Time} & \multirow{2}{*}{Max Batch Size} \\
& Query/Doc \\
\midrule
$N=1$  &  152.43/141.90   & 68.45  & 128  \\
$N=2$  &  160.86/156.76   & 69.51  & 128  \\
$N=3$  &  174.91/181.53   & 70.61  & 128  \\
$N=4$  &  194.58/216.21   & 71.94  & 128  \\
$N=6$  &  250.78/315.29   & 76.20  & 128  \\
$N=8$  &  329.46/454.00   & 85.27  & 64   \\
$N=12$ &  554.26/850.32   & 105.39 & 32   \\
$N=24$ &  1768.18/2770.74 & 187.03 & 16   \\
\bottomrule
\end{tabular}}
\caption{
Inference efficiency of different token compression settings, measured by 1000 randomly selected testset pairs.
Models are accelerated by FlashAttention-2 in float16.
$N=24$ is equivalent to the DPR baseline.
} \label{tab:efficiency}
\end{table}

\paragraph{RQ5: Encoding efficiency of MME. Tab. \ref{tab:efficiency}}
The Matryoshka-style visual token also brings an enhancement in encoding efficiency, reducing the computational overhead of the large LLM backbone \cite{cai2024matryoshka}.
To quantify the gain, we randomly select 1000 query-document pairs from the testset and measure the average sequence length, encoding time, and maximum batch size for different settings.
Table \ref{tab:efficiency} shows the results.
In our MME (\S\ref{sec:method:mme}), the visual token size of each image is controlled by the grid width $N$.
As expected, decreasing $N$ leads to reduced visual token numbers (sequence length), which translates into both accelerated encoding speeds (shorter time) and enhanced batch processing capabilities (larger batch size).
In practice, the optimal $N$ is determined by the trade-off between encoding efficiency and retrieval performance (Figure \ref{fig:token-curve}), which allows for flexible and efficient model deployment.

\section{Related Work}

\subsection{Multimodal Information Retrieval}
Early Multimodal Information Retrieval tasks focused on cross-modal retrieval of text and image~\citep{DBLP:conf/ijcai/CaoLLNZ22}, where the goal is simply to retrieve captions of everyday images \cite{lin2014microsoft,young-etal-2014-image}.
The scope has been extended to more complex scenarios, such as composed image retrieval \cite{liu2021image}, scientific contents \cite{wu-etal-2024-scimmir}, and visual documents \cite{ma-etal-2024-unifying,faysse2024colpali}.
Recent studies have been progressively exploring unified MIR settings \cite{zhou-etal-2024-marvel}.
For instance, M-BEIR \cite{DBLP:conf/eccv/WeiCCHZFRC24} integrates various image and text-related retrieval tasks, while UMRB \cite{zhang2024gme} further extends the evaluation to encompass more textual datasets and visual document retrieval \cite{faysse2024colpali}.
However, these benchmarks are constrained by their limitation to single-image queries or texts \cite{zhang2024gme}, lacking support for multi-image and interleaved contents.
We construct a new text-image interleaved retrieval benchmark to meet the demands of complex multimodal RAG scenarios.

Current strong multimodal retrievers predominantly adopt the dense retrieval paradigm, which can be categorized into two main approaches: CLIP-style dual-stream models \cite{liu2023universal,koukounas2024jinaclipv2,nussbaum2024nomic} and language model-centric architectures \cite{lin-etal-2024-preflmr,zhou-etal-2024-vista,jiang2024e5}.
\citet{wang-etal-2024-unified} proposed to compute unified multimodal embeddings from frozen LLM, which is not specifically designed for TIIR but shows potential in the multimodal context to image search task.
A concurrent work \cite{lee2024unified} also explores interleaved embeddings for multimodal document retrieval, where a task-specific hierarchical encoder is suggested to retrieve interleaved documents parsed from Wikipedia webpage.
In this work, we introduce the more generalized MLLM-based embedding model and propose a novel Matryoshka Multimodal Embedder to address the challenge of excessive visual tokens, which is crucial for TIIR.

\subsection{Multimodal Interleaved Modeling}
The modeling of interleaved text and image has been explored in various aspects, such as pre-training models \cite{NEURIPS2022_960a172b,laurenccon2024building} and corpus \cite{laurenccon2023obelics,zhu2023multimodal}.
Notably, there exists a parallel line of research focusing on unified models that simultaneously handle both interleaved representation and generation tasks \cite{pmlr-v202-koh23a,li2024improving,zou2024interfacing}.
Their experimental datasets are converted from existing multimodal generation datasets with interleaved context, \eg Visual Storytelling \cite{huang-etal-2016-visual}, and less retrieval-oriented.
Additionally, general interleaved corpus typically suffers from low knowledge density and logical coherence in image
sequence \cite{zhang20252}, which might not be suitable for constructing an interleaved retrieval benchmark.
In contrast, we build the TIIR dataset from human-curated high-quality tutorials (from wikiHow) for everyday skills, which are naturally interleaved and more informative for retrieval.

\section{Conclusion}
In this work, we introduce a new Text-Image Interleaved Retrieval (TIIR) task where the query and document are interleaved sequences of text and images, requiring the multimodal retriever to understand the semantics from interleaved context.
We construct the wikiHow-TIIR benchmark based on the high-quality tutorial corpus from wikiHow, and present an efficient pipeline to generate text-image interleaved queries.
We adapt several non-interleaved off-the-shelf multimodal and text retrievers to evaluate on our benchmark, showing that keeping interleaved structure is crucial for TIIR modeling.
To explore native interleaved retrievers, we train interleaved MLLM-based DPR baseline and propose a novel Matryoshka Multimodal Embedder (MME) to address the challenge of excessive visual tokens.
Evaluation results demonstrate the visual token compression strategy of MME achieves better performance and efficiency.
We also present extensive analyses to understand the TIIR task and models, providing insights for future research in multimodal retrieval.


\bibliography{custom}


\appendix

\section*{Appendix}

\section{WikiHow-TIIR}\label{sec:app:data}

\subsection{Data Collection}\label{sec:app:data:collect}

Our corpus construction adopts the wikiHow articles collected by  \citet{yang-etal-2021-visual}, systematically curated for Visual Goal-Step Inference (VGSI) research. This dataset comprises approximately 53,000 instructional articles. Structurally, each article decomposes a procedural objective (e.g., "hanging an ironing board") into multiple implementation methods (each article contains an average of 3 methods), with every method further annotated as stepwise components containing: 
(1) step headlines, 
(2) detailed descriptions, and 
(3) corresponding image demonstrations.
We convert them into 155,262 self-contained, text-image interleaved documents, each structured as <Goal, Method Name, [(Step-Headline, Step-Image), ...]>.

Our multimodal query generation pipeline employs three state-of-the-art open-source architectures: Idefics3-8B-Llama3 \cite{laurenccon2024building}, Qwen2.5-72B-Instruct \cite{qwen2}, and FLUX.1-dev \cite{flux2023}.  The workflow initiates with systematic extraction of categorical metadata from wikiHow, successfully curating annotations for around 29,000 articles.  Through stratified random sampling constrained by category distribution, we constructed: (1) A human-annotated test corpus comprising 7,654 queries and (2) A sample training partition containing 25,000 articles (pairs=73,084).

\begin{figure}[h]\centering
\includegraphics[width=\linewidth]{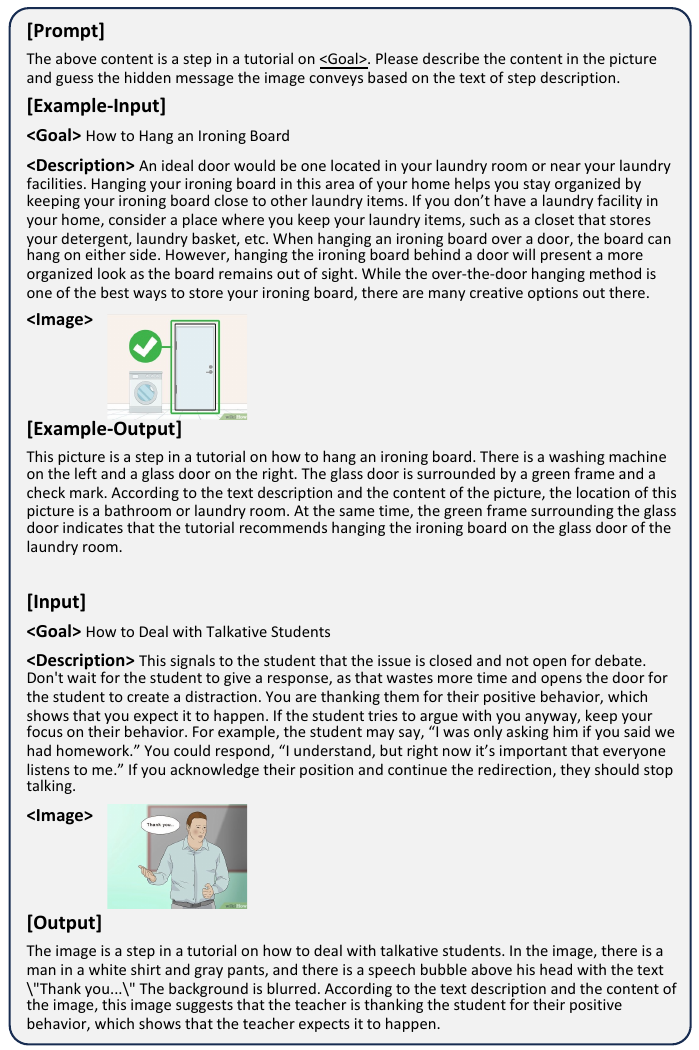}
\caption{
The example of the prompt, input and output of image caption.
}\label{fig:gen-cap-example}
\end{figure}

\subsection{Query Generation}\label{sec:app:data:query-gen}
\label{subsec:Query-Generation}
\subsubsection{Query Text Generation}
The reason why we select LLM to generate textual queries instead of MLLM is that:
\textbf{(1)} At the time we conduct the study, MLLMs are not powerful enough to accept text-image interleaved data to perform complex task generation.
\textbf{(2)} Considering that we add design examples to the data generation process, if we use MLLM, we need to input more than ten images at a time, or even more, which brings great challenges to machine performance, runtime, and model capability.
\textbf{(3)} Describe the image in the document through MLLM first and then use the textual document to generate data through LLM can effectively use the powerful performance of the current LLM, and can get better data generation effect in less resources and shorter running time.

\begin{figure}[h]\centering
\includegraphics[width=\linewidth]{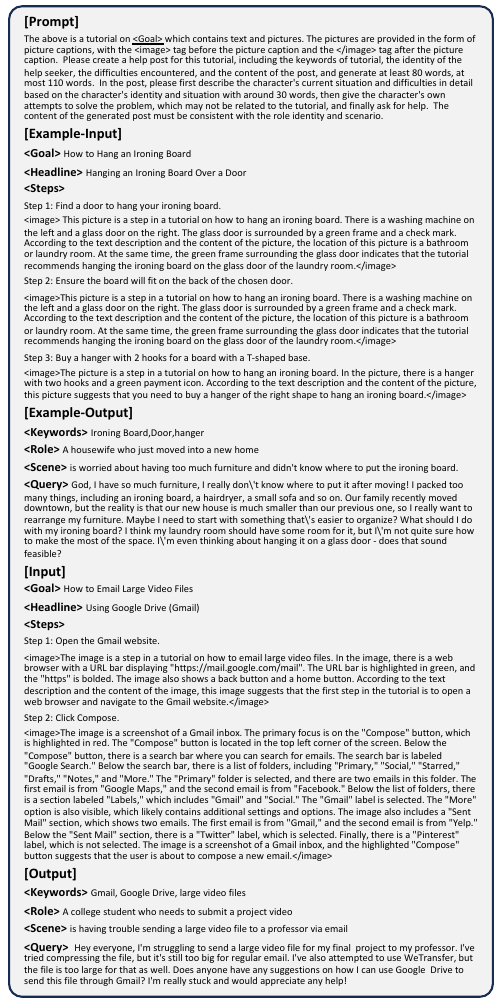}
\caption{
The example of the prompt, input and output of Query Text Generation.
}\label{fig:text-query-gen-example}
\end{figure}

\paragraph{Image Caption}
Therefore, we convert images to textual descriptions using Idefics3 by in-context learning style prompting. We chose this model considering that we fill in a well-designed example and the need to strike a balance between interleaved cross-modal alignment accuracy and computational efficiency. Specifically, we decompose each method into discrete steps and sequentially input stepwise data into the model to generate image captions that extract latent visual semantics. The implementation example is illustrated in Figure \ref{fig:gen-cap-example}.

\begin{figure}\centering
\includegraphics[width=\linewidth]{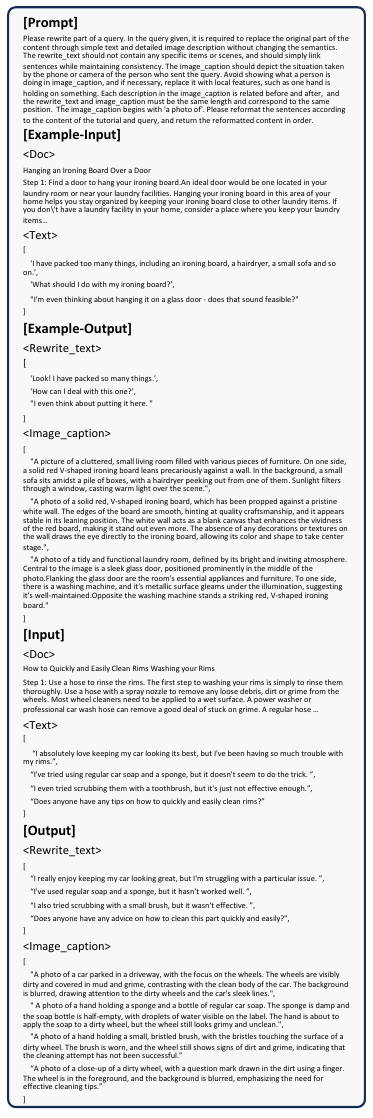}
\caption{
The example of the prompt, input and output of Text-image Information Reorganization.
}\label{fig:rewrite_query_example}
\end{figure}

\paragraph{Query Text Generation}
Following the text-only conversion of interleaved multimodal documents, we implement a two-stage query generation pipeline using a LLM. Current MLLMs (e.g., Chameleon \cite{Chameleon_Team_Chameleon_Mixed-Modal_Early-Fusion_2024}) with joint text-image generation capabilities lack accessible image generation modules, necessitating sequential construction of image-text interleaved queries through: 
(1) Primary textual query synthesis using Qwen2.5-72B-Instruct, and 
(2) Subsequent multimodal composition. 
The Qwen2.5-72B-Instruct architecture is configured with a multi-perspective prompting framework across four semantic axes: keywords, character, scene, and query, simulating real-world problem-solving scenarios. The implementation example is demonstrated in Figure \ref{fig:text-query-gen-example}.

\subsubsection{Text-image Information Reorganization}
The construction of text-image interleaved queries presents dual modality coordination challenges during partial textual substitution:
First, naive text-to-image conversion without original text retention induces inter-modal incoherence, where visual outputs fail to maintain linguistic continuity. Concurrently, directly processing non-objective textual queries through image generation models leads to visual semantic ambiguity due to conceptual abstraction. 
Second, preserving original textual components risks semantic redundancy, where visual representations become subsumed by textual semantics, negating their informational value.
To solve these problems, we  identify substitutable textual segments through semantic saliency analysis.

We implement a two-phase optimization method:
Phase 1: Visual Info Selection. we segment query texts into constituent sentences and perform relevance ranking against source documents using BM25 to isolate the top-k ($k={2,3,4}$) maximally informative sentences.
Phase 2: Query Rewriting. The selected sentences undergo semantic transformation via Qwen2.5-72B-Instruct, which:
(1) Simulates human multimodal communication patterns by substituting text narratives with visual representations. 
(2) Synthesizes contextual bridging statements to maintain discourse continuity.
This dual phase approach ensures the preservation of informational fidelity while achieving a human-aligned modality distribution, as demonstrated in Figure \ref{fig:rewrite_query_example}.

\subsubsection{Image Generation}
\begin{figure}[t]\centering
\includegraphics[width=\linewidth]{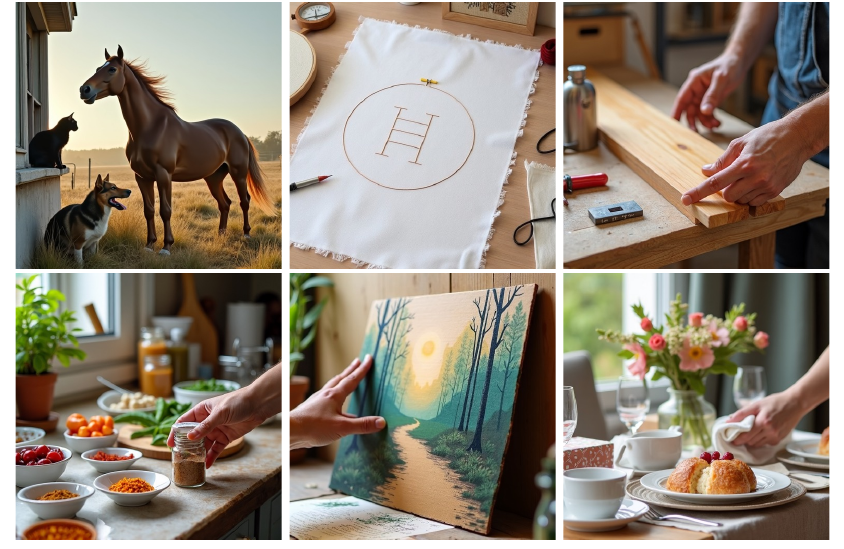}
\caption{
Examples of generated images.
}\label{fig:gen_image-example}
\end{figure}

The image generation phase employs FLUX.1-dev, a state-of-the-art open-source image generation model, to generate images from captions. We configure the model with photorealistic constraints through the prompt ["photorealistic", "realistic", "photograph"] and set the output resolution to 512×512 pixels to ensure spatial consistency. The generated images are illustrated in Figure \ref{fig:gen_image-example}.

\begin{figure}[t]\centering
\includegraphics[width=\linewidth]{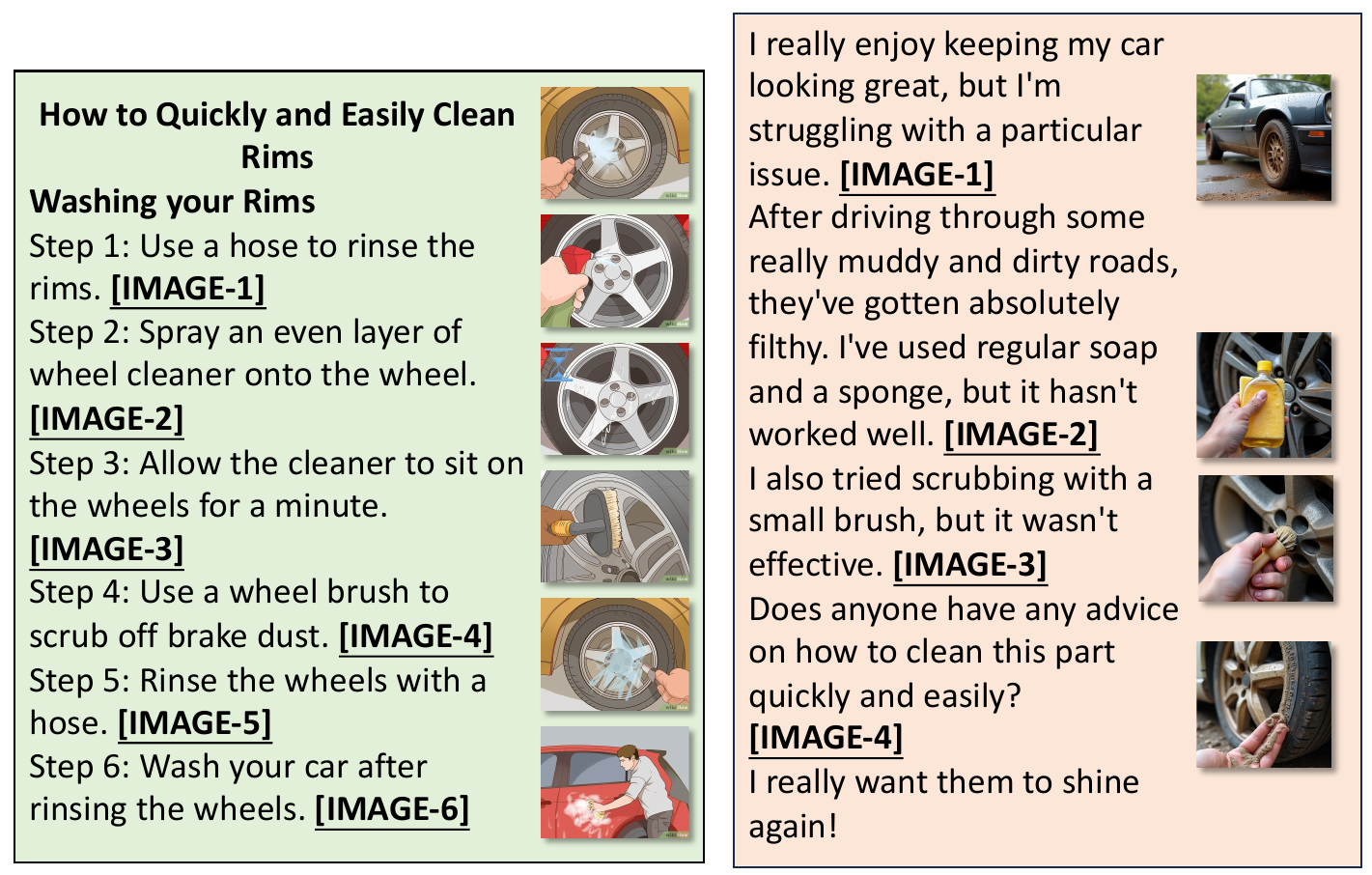}
\caption{
The example of our WikiHow-TIIR document and query.
}\label{fig:querydoc-example}
\end{figure}

\subsection{Data Annotation}\label{sec:app:data:annotation}
We deploy a web-based annotation interface using Label Studio \cite{Label_Studio}, hosting around 10,000 test instances requiring labeling, and engage 10 computer science graduate annotators via the university's information platform. After annotation, we implement dual verification mechanisms that include random sampling and statistical consistency checks.  Annotators received performance-based remuneration calculated with hourly compensation rates averaging 12\$, exceeding local academic compensation standards.

On the whole, we establish strict guidelines that prioritize ethical and safety considerations, requiring all queries to: 
(1) adhere to legal standards, 
(2) exclude content involving pornography, violence or illegal activities, and 
(3) demonstrate rational and contextually appropriate requests.

We design an annotation methodology for image annotation comprising three key assessment dimensions:
(1) Structural Integrity Evaluation: Annotators identify morphological anomalies in character and object generation.
(2) Textual Content Classification: A three-tier text quality assessment. Level 1: No text. Level 2: Legible and comprehensible text. Level 3: Obvious textual errors
(3) Semantic Relevance Verification. Annotators determine the image's contextual meaningfulness, excluding instances unrelated to the query or document.

Moreover, we set a comprehensive coherence evaluation methodology to address potential inconsistencies arising from independent image generation:
Level 1: Consistent subject/scene representation.
Level 2: Minimal variations in subject/scene characteristics.
Level 3: Significant divergences in subject/scene depiction.
Annotators holistically analyze all images within a single query, systematically assessing visual consistency and identifying potential generative model limitations in maintaining semantic and visual coherence.

\subsection{Data Statistics}
Table \ref{tab:data-stats} presents the dataset statistics. We calculate average text token lengths by concatenating text chunks and encoding them using LlamaTokenizer. Following the query generation methodology in \S\ref{subsec:Query-Generation}, we create one positive query per document while utilizing same-article documents as hard negative samples (as stated in \S\ref{sec:app:data:collect}, each article contains an average of three documents).

\begin{figure}[t]\centering
\includegraphics[width=\linewidth]{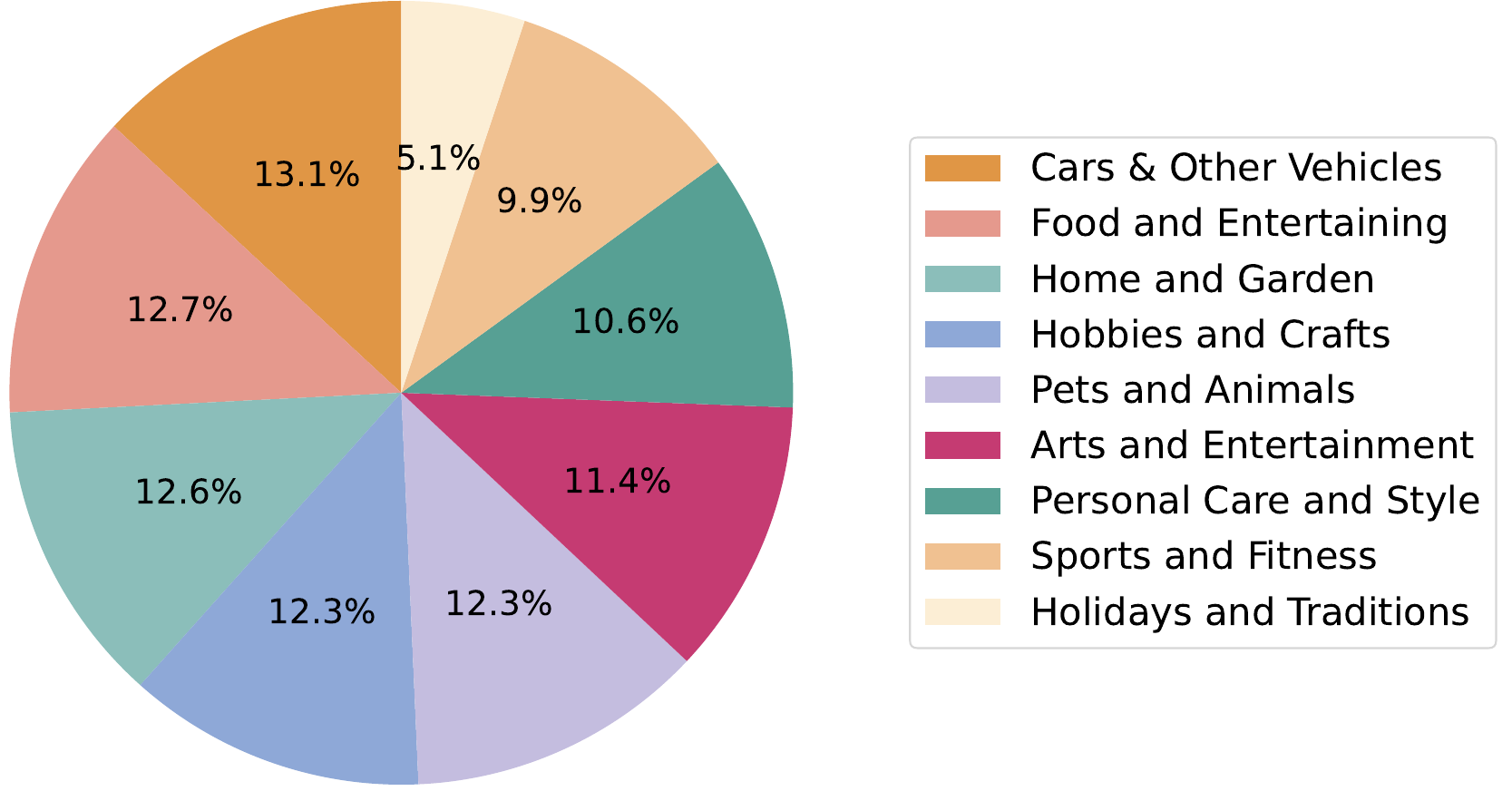}
\caption{
Categories of test dataset.
}\label{fig:test-category}
\end{figure}

Figure \ref{fig:test-category} illustrates the category distribution in our test set, which covers nine real-life domains: Vehicles, Food, Home Improvement, Crafts, Animals, Arts, Personal Care, Fitness, and Traditions. Sourced from wikiHow articles, this categorization comprehensively represents common human activities, demonstrating the test set's representativeness for fair evaluation.

\section{Implementation Details}\label{sec:app:implementation}
We fine-tune OpenAI CLIP and DeepSeek-VL-1.3B. During training, we use a batch size of 32 and set a learning rate of $5 \times 10^{-5}$/$2 \times 10^{-5}$ with a linear warm-up scheduler for DeepSeek-VL-1.3B/CLIP. In our contrastive learning configuration, the temperature coefficient $\tau$ is empirically set to 0.05. Documents derived from identical source articles are designated as in-batch negatives. Specifically, we implement randomized selection of a single hard negative instance per mini-batch. The entire process undergoes three complete training epochs.

We select DeepSeek-VL-1.3B-base to train in four ways. 
(1) \textit{Baseline}(DPR): We set the image token number as the model default, 576, to train. 
(2) \textit{Random sampling} (Rand):We randomly sample a grid width $N$ for each micro-batch.
(3) \textit{Matryoshka learning} (MRL): We train the model with all $M$ kernel sizes simultaneously.
(4) \textit{Mean learning} (Mean): We additionally compute losses between query and document embeddings of different sizes, the final loss is the mean of all $M \times M$ possible combinations. 
All models are trained with the max token length of 4096, and test with the same.

Table \ref{tab:clip-result} demonstrates Jina-CLIP-v2's superior performance through  normalized image-text embedding fusion approach (summation of averaged modality embeddings). This methodology was subsequently adopted for training clip-vit-large-patch14, with detailed performance metrics provided in the same table.

\section{Experiments Details}
All experiments are conducted on a NVIDIA A100-80G 8-GPU server.
All retrieval results were implemented using Faiss \cite{douze2024faiss}.

\subsection{Single-image Multimodal Retrievers}
\begin{figure}[t]\centering
\includegraphics[width=\linewidth]{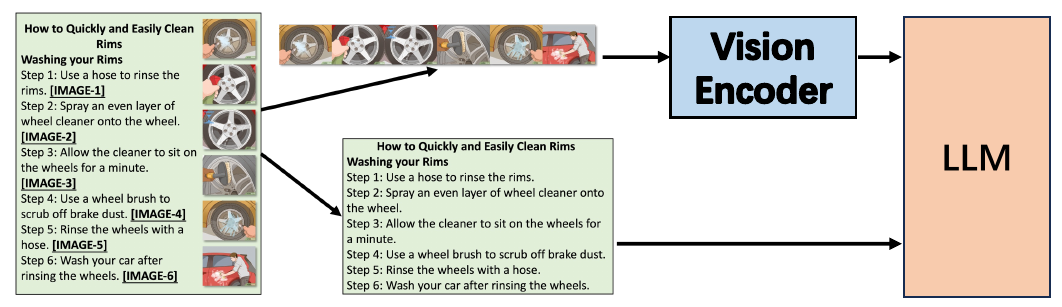}
\caption{
The example of the way that encodes text-image interleaved content with single-image multimodal retrievers.
}\label{fig:single-image-example}
\end{figure}

Given architectural constraints in single-image multimodal retrievalers that process only single image-text pairs per instance, we disentangle image-text interleaved data into images and text to encode. The implementation pipeline (Figure \ref{fig:single-image-example}) demonstrates this separation process.

E5-V introduces unimodal training through text-only pairwise optimization. The architecture employs specialized markup templates for modality-specific encoding. The constructed prompts what we set are formally specified in Table \ref{tab:E5V-prompt} following standard template formatting conventions.

\begin{table}
\resizebox{\linewidth}{!}{
\setlength{\tabcolsep}{4pt}
\begin{tabular}{ll}
\toprule
Format & Prompt(query/doc) \\
\midrule
only-text & <sent>\textbackslash{}nSummary above query/tutorial in one word: \\
image+text & <image>\textbackslash{}n<sent>\textbackslash{}nSummary above query/tutorial in one word: \\
\bottomrule
\end{tabular}}
\caption{The instructions of the E5-V model.}
\label{tab:E5V-prompt}
\end{table}

MM-Embed and GME$_\text{Qwen2-VL-2B}$ require task-specific instructions appended to each query. We implement standardized prompts for both architectures: "Retrieve a wikiHow tutorial that provides an answer to the given query" for MM-Embed and "Find a wikiHow tutorial that matches the given query" for GME$_\text{Qwen2-VL-2B}$.

\subsection{Text Models}\label{sec:app:text-model}
For text models, we implement two encoding strategies for text-image interleaved data: (1) remove the images and keep only the text, and (2) replace the images with image captions. The latter employs the standardized prompt "Describe the image" for real-time inference simulation, replacing image with generated captions through the processing of Qwen2-VL-2B-Instruct.

We implement standardized prompt "Given a query, retrieve relevant wikiHow document that answer the query" for GTE-Qwen2-7B and "Represent this query for searching relevant wikiHow passages:" for BGE-v1.5$_\text{large}$.

\subsection{Two-stream Models}
For two-stream models, we employ separate text-image encoding pipelines. Text embeddings derive from concatenated document chunks, while visual encoding explores: 
(1) image concatenation, and 
(2) normalized mean pooling of individual image embeddings. 
Following established multimodal fusion methods \cite{liu2023universal}, we evaluate three combination strategies: vector summation, feature concatenation, and element-wise multiplication, reporting optimal results in Table \ref{tab:main}.

\subsection{Visual Document (Image) Retrievers}
\begin{figure}[t]\centering
\includegraphics[width=\linewidth]{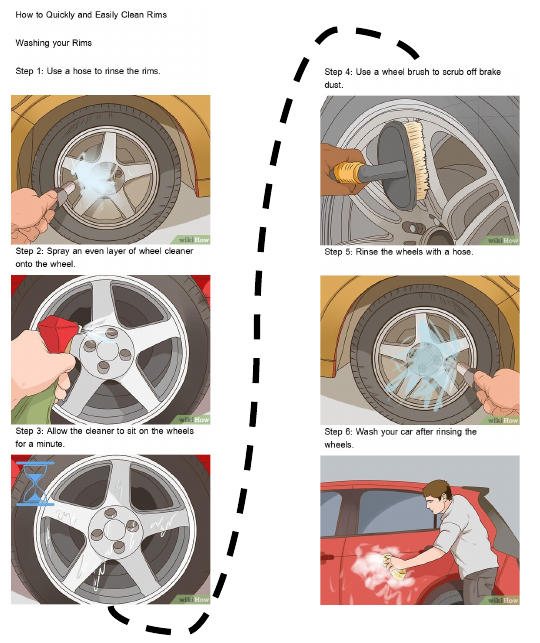}
\caption{
The example of visual document (image). The left and right images in the picture are joined up and down, but for the sake of the layout of the paper, we cut them and arrange them left and right.
}\label{fig:text2image}
\end{figure}

For visual document (image) retrievers, we convert the whole query/document into one image. The example is shown in Figure \ref{fig:text2image}.

\begin{table}
\resizebox{\linewidth}{!}{
\setlength{\tabcolsep}{4pt}
\begin{tabular}{l|ll|ccc}
\hline
Model & Text\&Image & Image & Recall@5 & MRR@10 & nDCG@10 \\ 
\hline
\multirow{6}*{Jina-CLIP-v2 }  & \multirow{2}*{Sum}  & mean & 58.80 & 45.00 & 47.17  \\ 
  & ~ & concat & 51.13 & 38.30 & 40.22  \\ 
  & \multirow{2}*{Concatenate}  & mean & 55.91 & 43.28 & 45.25  \\ 
  & ~ & concat & 50.10 & 37.35 & 39.37  \\ 
  & \multirow{2}*{Dot product}  & mean & 30.36 & 22.04 & 23.14  \\ 
~ & ~ & concat & 24.61 & 17.80 & 18.61  \\ 
\hline
\multirow{6}*{
\begin{tabular}{c} CLIP$_\text{large}$ \\ Fine-tuned \end{tabular}
}  & \multirow{2}*{Sum}  & mean & 69.41 & 54.73 & 57.15  \\ 
  & ~ & concat & 55.55 & 42.27 & 44.18  \\ 
  & \multirow{2}*{Concatenate}  & mean & 61.18 & 47.4 & 49.55  \\ 
  & ~ & concat & 49.33 & 37.34 & 38.9  \\ 
~ & \multirow{2}*{Dot product}   & mean & 16.19 & 12.09 & 12.45  \\ 
~ & ~ & concat & 10.5 & 7.64 & 7.92 \\ 
\hline
\end{tabular}}
\caption{Evaluation results on our WikiHow TIIR of the two-stream models, Text\&Image denotes the way we combine the text and image embedding, and Image denotes the way we get the image embedding.}
\label{tab:clip-result}
\end{table}

\subsection{Ablation Study}
Finally, we conduct an ablation study to investigate the hyper-parameters in our model training.
Due to computational constraints\footnote{
The training instances of our dataset frequently generate input sequences with lengths in the order of 4,000 tokens, resulting in substantial memory consumption.
}, our hyper-parameter search is based-on the most training-friendly \texttt{Rand} strategy of MME.
We vary the rank of LoRA (8, 16, 32) and learning rate (1e-4, 2e-5), where the LoRA rank controls the size of new learnable parameters in training.
Although batch size substantially influences model performance (with larger batch sizes generally yielding better results in contrastive learning), we opt to maintain a fixed batch size, \ie the maximum allowable within GPU constraints, across all models to ensure fair comparison.
Therefore, the impact of batch size is not discussed in this analysis.
As shown in Table \ref{tab:ablation}, the best setting is achieved with a rank of 16 and a learning rate of 5e-5.

\begin{table}
\resizebox{\linewidth}{!}{
\setlength{\tabcolsep}{4pt}
\begin{tabular}{cccc}
\toprule
Model & LoRA Rank & Learning Rate & MRR@10 ($N=3$) \\
\midrule
\multirow{4}{*}{
\begin{tabular}{c} MME \\ Rand \end{tabular}
}
& 16 & 5e-5 & 62.89 \\ 
& 16 & 1e-4 & 58.18 \\ 
& 8  & 5e-5 & 62.52 \\ 
& 32 & 5e-5 & 62.36 \\ 
\bottomrule
\end{tabular}}
\caption{
Ablation study of different hyper-parameters in our MLLM-base model training.
We perform hyper-parameter search on MME align since it's the fastest to train.
The results of the best setting $N=3$ are shown.
As GPU resources are limited, we run all experiments with the same batch size of 32.
} \label{tab:ablation}
\end{table}


\end{document}